\documentclass[10pt,twocolumn,letterpaper]{article}

\usepackage{cvpr}              
\usepackage[accsupp]{axessibility}

\usepackage[dvipsnames]{xcolor}

\graphicspath{{figures/}}

\usepackage{pifont}
\usepackage{booktabs}
\usepackage{multirow}
\usepackage{rotating}
\usepackage{algorithm}
\usepackage{algorithmic}
\usepackage{capt-of}
\usepackage{url}
\usepackage{mathtools}
\usepackage{subcaption}
\usepackage{colortbl}

\newcommand{\Tref}[1]{Table~\ref{#1}}

\newcommand{\Fref}[1]{Fig.~\ref{#1}}

\newcommand{\Sref}[1]{Section~\ref{#1}}

\newcommand{\cmark}{\ding{51}}%
\newcommand{\xmark}{\ding{55}}%

\definecolor{cvprblue}{rgb}{0.21,0.49,0.74}
\usepackage[pagebackref,breaklinks,colorlinks,citecolor=cvprblue]{hyperref}

\def\paperID{2488} 
\def\confName{CVPR}
\def\confYear{2024}

\title{Exploring Vision Transformers for 3D Human Motion-Language Models \\ with Motion Patches}

\author{Qing Yu  \ \ \ \ Mikihiro Tanaka \ \ \ \  Kent Fujiwara \\
LY Corporation \\
{\tt\small \{yu.qing,~mikihiro.tanaka,~kent.fujiwara\}@lycorp.co.jp}
}

\begin{document}
\maketitle

\begin{abstract}
To build a cross-modal latent space between 3D human motion and language, acquiring large-scale and high-quality human motion data is crucial. However, unlike the abundance of image data, the scarcity of motion data has limited the performance of existing motion-language models. To counter this, we introduce ``motion patches'', a new representation of motion sequences, and propose using Vision Transformers (ViT) as motion encoders via transfer learning, aiming to extract useful knowledge from the image domain and apply it to the motion domain. These motion patches, created by dividing and sorting skeleton joints based on body parts in motion sequences, are robust to varying skeleton structures, and can be regarded as color image patches in ViT. We find that transfer learning with pre-trained weights of ViT obtained through training with 2D image data can boost the performance of motion analysis, presenting a promising direction for addressing the issue of limited motion data. Our extensive experiments show that the proposed motion patches, used jointly with ViT, achieve state-of-the-art performance in the benchmarks of text-to-motion retrieval, and other novel challenging tasks, such as cross-skeleton recognition, zero-shot motion classification, and human interaction recognition, which are currently impeded by the lack of data.
\end{abstract}    
\section{Introduction}
\label{sec:intro}

The cross-modal analysis of 3D human motion and natural language has opened up new avenues for tasks such as motion recognition~\cite{vemulapalli2014human, du2015hierarchical, tang2018deep, yan2018spatial, shi2019two, petrovich2023tmr, tevet2022motionclip, zhu2023motionbert} and text-to-motion synthesis~\cite{athanasiou2022teach, petrovich2022temos, petrovich2021action}, which can benefit the applications like animating avatars or humans \cite{xu2023magicanimate, hu2023animate}. The key to these tasks is constructing a cross-modal latent space that captures the intricate relationship between human motions and language semantics, allowing systems to interpret and generate human-like motions based on textual descriptions. 

\begin{figure}[t]
    \centering
    \includegraphics[width=1\linewidth]{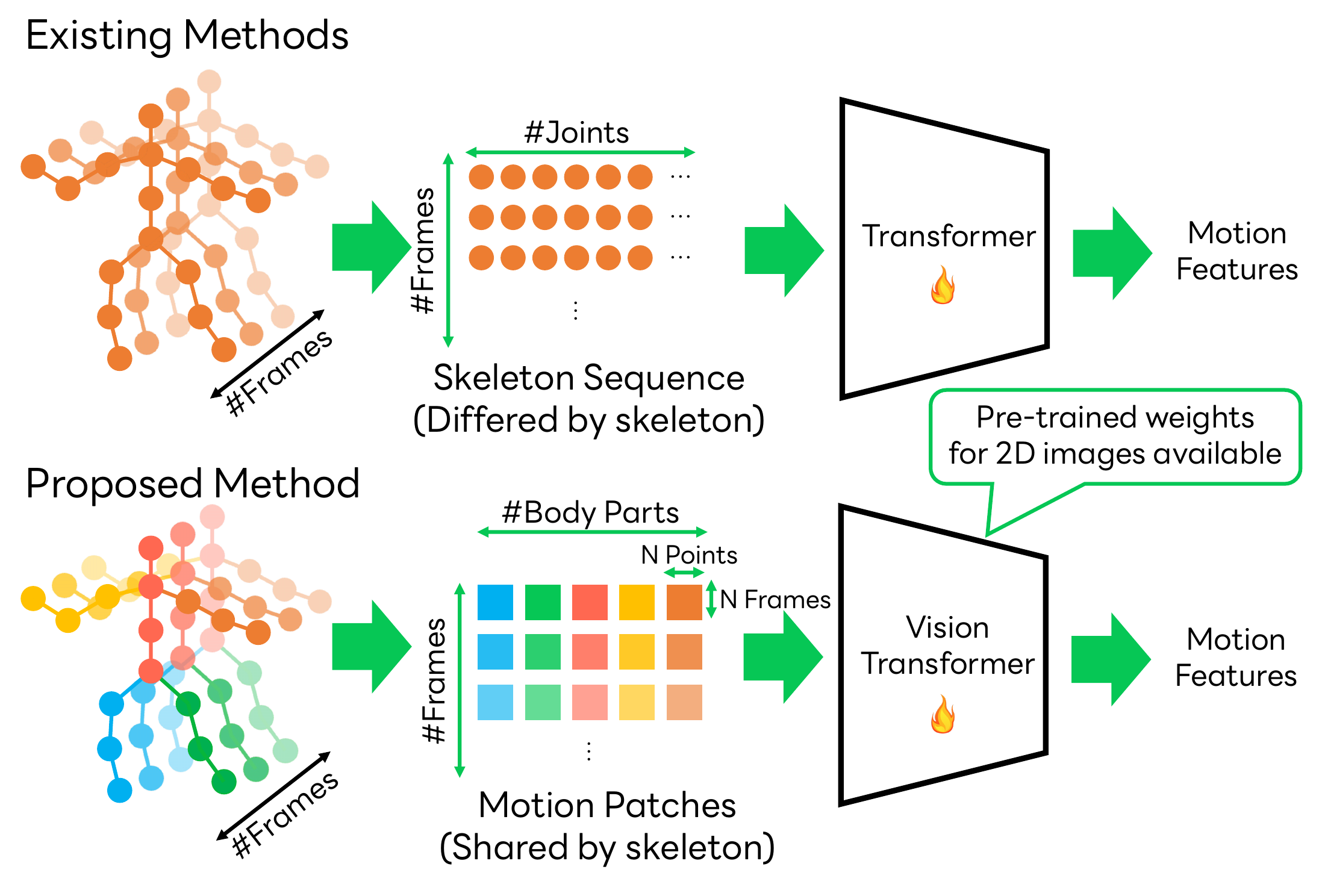}
    \vspace{-15pt}
    \caption{Overview of the existing methods and the proposed method. The existing methods train an original Transformer with the joint information from the motion sequences directly, while the proposed method converts them into motion patches and then trains the ViT, which can be initialized with pre-trained weights.}
    \vspace{-5pt}
    \label{fig:setting}
\end{figure}

Despite the promising advancements in this area, one of the most challenging aspects is the scarcity of data, because the process of collecting and annotating 3D human motion data is labor-intensive and time-consuming. While recent years have seen an increase of motion capture datasets~\cite{mahmood2019amass}, some of which are categorized by classes~\cite{liu2019ntu} or even labeled with free text~\cite{punnakkal2021babel, plappert2016kit}, these resources are still not sufficient for deep learning algorithms to fully understand human motions. Moreover, because various motion capture systems and skeleton structures are used in different datasets, it has been difficult to build a large-scale dataset with a unified representation.

To build a motion-language model, existing methods~\cite{petrovich2023tmr, tevet2022motionclip} attempt to incorporate text embeddings into motion autoencoders. Due to the lack of large-scale data, they train the motion encoder from scratch on each dataset and try to use motion synthesis in autoencoders to improve the motion features. Tevet et al.~\cite{tevet2022motionclip} attempt to apply pre-trained image-language models~\cite{radford2021learning} to motion data, but they only render a single frame as the input image. Consequently, these methods are not yet robust enough to handle the variations and subtleties present in 3D human motions. They also need to deal with different skeleton structures by training an individual model for each dataset separately, despite all these data representing human motion. This leads to low performance on the small-scale datasets.

To overcome these challenges, we introduce an approach to building motion-language models by leveraging Vision Transformers (ViT)~\cite{dosovitskiy2020image} as the motion encoder. This approach extends the conventional use of ViTs, which were originally designed for 2D image classification, to the more complex domain of 3D human motion analysis. With the transfer learning of ViT pretrained on ImageNet~\cite{russakovsky2015imagenet} to motion data, the training process of the motion encoder can be accelerated, while at the same time, overcoming the issue of limited data scale and achieving better correspondence between the motion features and the language features. 

To efficiently transfer knowledge of the image domain to the motion domain, we also propose ``motion patches'', a novel unified representation for various skeleton structures in motion sequences. We design the motion patches to be likened to image patches in ViT, with joint positions in $xyz$ coordinates simply converted to image colors in $rgb$ space. We first partition the joints of the skeleton into five body parts: torso, left arm, right arm, left leg, and right leg. Then, motion patches are formed by sampling $N$ points from each body part through linear interpolation, and stacking these points for each part across $N$ frames by sliding window. This results in a patch for each part with a size of $N \times N$. We then train a motion-language model with a contrastive learning framework~\cite{radford2021learning}. The comparison between existing methods and the proposed method is shown in~\Fref{fig:setting}.

We evaluate the versatility and effectiveness of our proposed method through comprehensive experiments and applications of motion-language tasks. This study makes the following contributions:
\begin{itemize}
    \setlength\itemsep{0pt}
    \item We propose a new framework for building motion-language models using ViT, which extends the application of ViT to obtain a cross-modal latent space between motions and language.
    
    \item We introduce a novel method of representing 3D human motion data as ``motion patches'', which can be processed by the ViT architecture with its pre-trained weights for transfer learning, and are also resilient to variations in human skeleton structures.

    \item Our approach not only significantly improves the performance of text-to-motion retrieval, but also illustrates the potential for other novel applications, such as cross-skeleton recognition, zero-shot motion classification, and human interaction recognition.
    
\end{itemize}
\section{Related Works}
\label{sec:related}

\begin{table}[t]
\begin{center}
\tabcolsep = 1.2mm
\resizebox{0.99\linewidth}{!}{
\begin{tabular}{c|c|c|c}
    \toprule
     Method & Input Type & Motion encoder & Unified Representation \\                 \midrule
     MotionCLIP~\cite{tevet2022motionclip} & \begin{tabular}{c}Motion \\+ Image\end{tabular}  & \begin{tabular}{c}Scratch Transformer \\+ Pre-trained CLIP\end{tabular} & \xmark \\
     TMR~\cite{petrovich2023tmr} & Motion  & Scratch Transformer & \xmark  \\ 
     Proposed & Motion Patch  & Pre-trained ViT & \cmark  \\
    \bottomrule
\end{tabular}
}
\end{center}
\vspace{-10pt}
\caption{Summary of recent related methods for motion-language models. Only our proposed method utilizes pre-trained motion encoders and a unified representation for various skeleton structures.}
\vspace{-5pt}
\label{tab:related}
\end{table}

\begin{figure*}[t]
    \centering
    \includegraphics[width=1\linewidth]{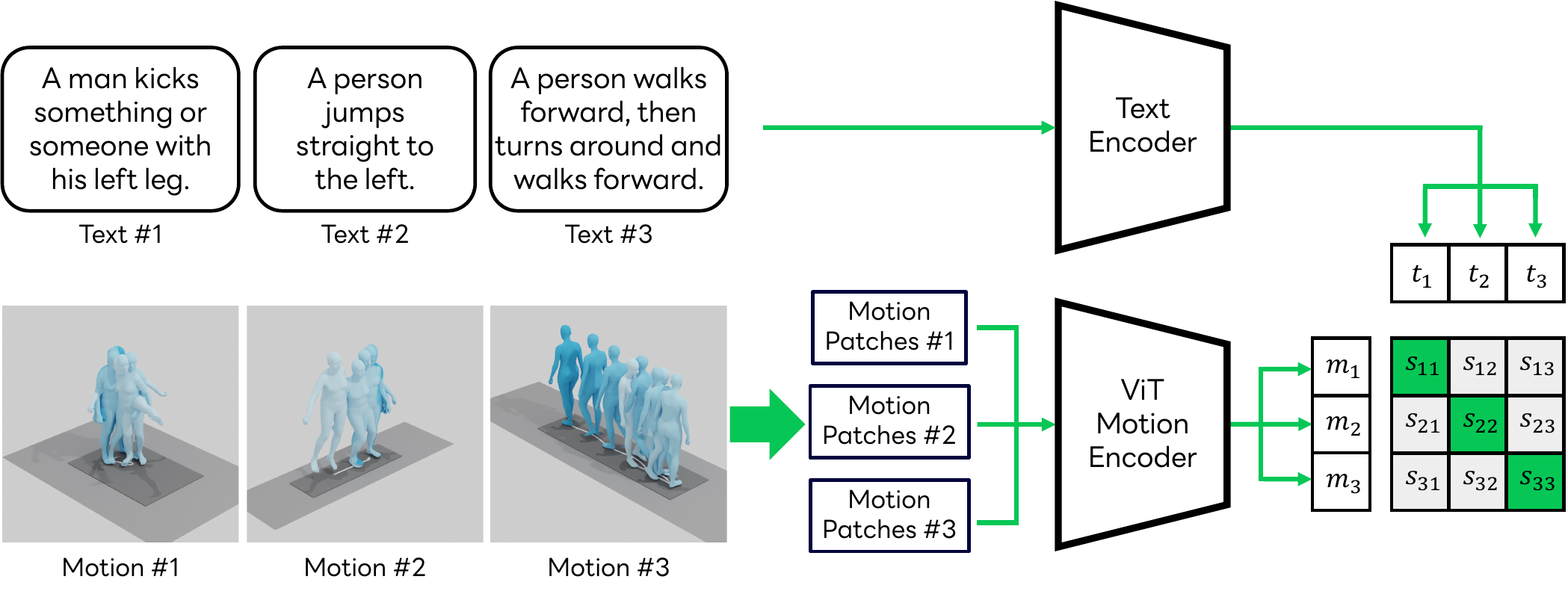}
    \vspace{-10pt}
    \caption{Overview of the proposed framework, which consists of a motion encoder and a text encoder. We transform the raw motion sequences into motion patches as the input of the ViT-based motion encoder. We calculate the similarity matrix between text-motion pairs within a batch to train the model. To illustrate this concept, we provide an example batch containing three samples for clarity.}
    \vspace{-5pt}
    \label{fig:method_overview}
\end{figure*}

\textbf{Motion-Language Datasets.}
While human motion modeling has gained interest in linking language and 3D body motions, there is a scarcity of motion-language datasets. Although datasets do exist for action recognition~\cite{shahroudy2016ntu} and pose estimation~\cite{ionescu2013human3}, they lack detailed textual descriptions for each motion. Notably, the KIT dataset~\cite{plappert2016kit} offers 11 hours of motion capture sequences, each paired with descriptive sentences. The recently released HumanML3D dataset~\cite{guo2022generating} provides around 29 hours of motion data with natural language labels for AMASS~\cite{mahmood2019amass} and HumanAct12~\cite{guo2020action2motion} collections. Compared to the scale of image-text pairs used for training image-language models like CLIP (\eg, datasets with 400 million images), motion-text pairs remain notably limited in scale (\eg, 14,616 motions in HumanML3D). Creating motion-language datasets presents challenges, including the need for expensive motion capture and annotation systems, as well as issues related to variations in skeleton structures across different datasets.

\textbf{Motion-Language Models.}
In recent years, vision-language foundation models have garnered significant attention, driven by the availability of vast collections of image-text pairs gathered from the internet. These models have adopted various pre-training schemes~\cite{lu2019vilbert, su2019vl, chen2019uniter, li2022blip}. A recent representative in this field is CLIP~\cite{radford2021learning}, which aims to learn joint representations of vision and language by training on a large-scale dataset of image-text pairs. However, the field of motion-language models is relatively less explored. Petrovich et al.~\cite{petrovich2023tmr} use contrastive training during motion generation to align text features with motion features. Because of the limited scale of motion-text pairs, these methods are trained from scratch and cannot capture the differences between similar motions. Moreover, these methods cannot be applied to cross-skeleton recognition, as different motion encoders need to be trained for each dataset. There are some attempts to utilize external knowledge from different modalities to analyze human motion. Tevet et al.~\cite{tevet2022motionclip} attempt to render a single frame as a static image to be used as input to CLIP, in order to obtain visual features and align them with motion features, but the performance is limited by the use of a single frame.

\textbf{Motion Generation and Retrieval.}
Motion-language models find valuable application in text-conditioned motion generation. Unlike unconstrained motion generation~\cite{yan2019convolutional, zhao2020bayesian}, action-conditioned~\cite{guo2020action2motion, petrovich2021action} or text-conditioned models~\cite{ahuja2019language2pose, athanasiou2022teach, chen2023executing, ghosh2021synthesis, kalakonda2022action, petrovich2022temos, tevet2022human, guo2022tm2t, zhang2023t2m, jiang2023motiongpt, tanaka2023role} introduce semantic controls to generate motion sequences corresponding to input textual descriptions. Recent advancements in text-to-image generation through diffusion generative models~\cite{dhariwal2021diffusion, rombach2022high} have led to methods such as~\cite{tevet2022human, zhang2022motiondiffuse, chen2023executing}, aiming to apply diffusion models to text-to-motion generation using the HumanML3D dataset, along with other approaches based on Large Language Models~\cite{zhang2023t2m, jiang2023motiongpt}. Text-to-motion retrieval~\cite{petrovich2023tmr} is an alternative and potentially complementary approach to generating motions corresponding to a given textual description, as a retrieval model can always return a realistic motion. Text-to-motion retrieval is also used as a tool to evaluate the performance of text-to-motion generation~\cite{guo2022generating, zhang2022motiondiffuse,dabral2023mofusion, jiang2023motiongpt}. We also use the task of text-to-motion retrieval as an important evaluation of the proposed method.

\textbf{Transfer Learning.}
Transfer learning involves taking models trained for a specific task using a large dataset, and extending their capabilities to address new tasks by leveraging prior knowledge to extract relevant features. Some examples of these attempts can be found in image segmentation~\cite{iglovikov2018ternausnet} and medical image analysis~\cite{Majkowska2019ChestRI, Gulshan2016DevelopmentAV}. Besides the image domain, transfer learning using ImageNet~\cite{russakovsky2015imagenet} pre-trained weights also performs well in video recognition~\cite{carreira2017quo} and even in audio classification~\cite{palanisamy2020rethinking, gong2021psla}, where time sequences are transformed into images as the input of the model. Some methods~\cite{laraba20173d, ali2023skeleton} attempt to use convolutional neural networks for action recognition. However, our focus is on leveraging the pre-trained knowledge from the image domain to construct motion-language models.

To address the challenges of motion-language models, we aim to transfer the ViT pre-trained in the image domain to the motion domain with a unified representation of motion sequences, to overcome the data scale problem. We summarize the differences between the proposed method and related motion-language models in~\Tref{tab:related}.

\section{Method}
\label{sec:method}

\subsection{Problem Statement}
\label{sec:problem}
Given a set of motion sequences $\mathcal{M}$ and a set of captions $\mathcal{T}$, our target is to learn a function $s(m_i, t_j)$ to calculate the similarity between the motion $m_i \in \mathcal{M}$ and the caption $t_j \in \mathcal{T}$. The objective of the $s(m_i, t_j)$ is to calculate a high similarity for relevant motion-text pairs and a low similarity score for irrelevant ones. The motion sequence $m_i \in \mathcal{M}$ is represented as a sequence of skeleton joints in this paper. Formally, the motion sequence is denoted by $m_i \in \mathbb{R}^{T \times J \times 3}$, where $T$ represents the length of the sequence, $J \times 3$ represents the position of the skeleton joints in Cartesian coordinates, ($x,y,z$).

To build a motion-language model, we adopt the CLIP framework~\cite{radford2021learning}, which consists of a motion encoder $\mathcal{F}_M$ and a language model $\mathcal{F}_T$. Using these encoders, we encode the motion sequence $m_i$ as $\mathcal{F}_M(m_i)$ and the caption $t_j$, and then calculate the similarity as follows:
\begin{equation}
    s(m_i, t_j) = \frac{\mathcal{F}_M(m_i)  \cdot \mathcal{F}_T(t_i)}{\lVert \mathcal{F}_M(m_i) \rVert \lVert \mathcal{F}_T(t_i) \rVert}.
\label{eq:cosine_sim}
\end{equation}
The overall architecture of the proposed framework is shown in~\Fref{fig:method_overview}.

\begin{figure}[t]
    \centering
    \includegraphics[width=1\linewidth]{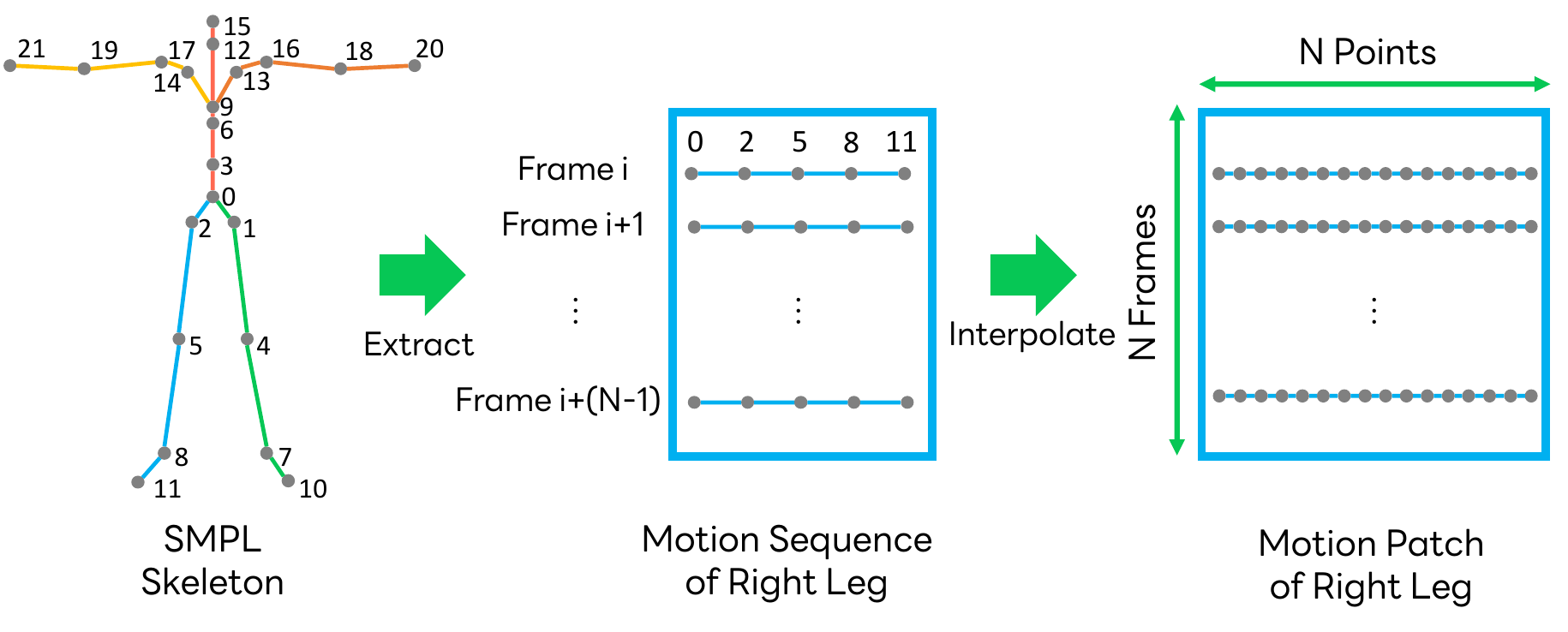}
    \vspace{-10pt}
    \caption{Process of building the motion patches for each motion sequence. Given a skeleton, we mark different body parts in different colors. We show the method to construct the motion patch of the right leg. The same process is applied to other body parts.}
    \label{fig:motion_patch}
    \vspace{-10pt}
\end{figure}

\subsection{Motion Patches}
\label{sec:motion_patches}
To extract spatial-temporal information in motion sequences as motion features, and to enable effective transfer of knowledge from the image domain via pre-trained models, motions need to be represented in a similar manner as images. However, in contrast to image data usually having a unified representation of size $224 \times 224 \times 3$ for deep learning models, the size of motion sequences is $T \times J \times D$ as mentioned in~\Sref{sec:problem}. Because the number of the frames $T$ differs for each sequence and the number of the joints $J$ depends on the skeleton structure in the dataset, there is no consensus on how to obtain a unified representation for motion data in different skeleton structures.

We propose ``motion patches'' as a new representation, which can be further used as input to ViT~\cite{dosovitskiy2020image} for motion feature extraction. To build motion patches similar to image patches with size $N \times N$ in ViT, we divide the joints of 3D motion skeletons into body parts, interpolate between joints in each body part to obtain $N$ sample points, and use the $N$ consecutive frames in the sequence as shown in~\Fref{fig:motion_patch}. First, the joints are partitioned into five body parts: the torso (including the head), left arm, right arm, left leg, and right leg. This type of partitioning is commonly used~\cite{jang2022motion} and can be implemented on any human skeleton. Each body part comprises a subset of joints corresponding to that part of the body according to the kinematic chain of the skeleton. 

Next, within each body part, we arrange the joints based on their distance from the torso. For example, in the case of arms, we order the joints as the upper chest $\rightarrow$ the shoulder $\rightarrow$ upper arm $\rightarrow$ lower arm $\rightarrow$ hand. This sequence maintains the spatial structure of the skeleton. We standardize the number of sample points in each body part to $N$ using linear interpolation. To normalize these sample points as image data, we calculate the mean and variance of each point across the dataset and perform the z-score normalization using these mean and variance values.

Finally, we form ``motion patches'' by stacking sequences of sample point positions across $N$ frames. We repeat this process for every sequence of $N$ frames using a sliding window, creating a series of motion patches. These patches, which are robust to variations in skeleton structures, can be analogized to image patches in ViT, and allow us to represent skeletons from various structures in a unified format.

\begin{figure}[t]
    \centering
    \includegraphics[width=1\linewidth]{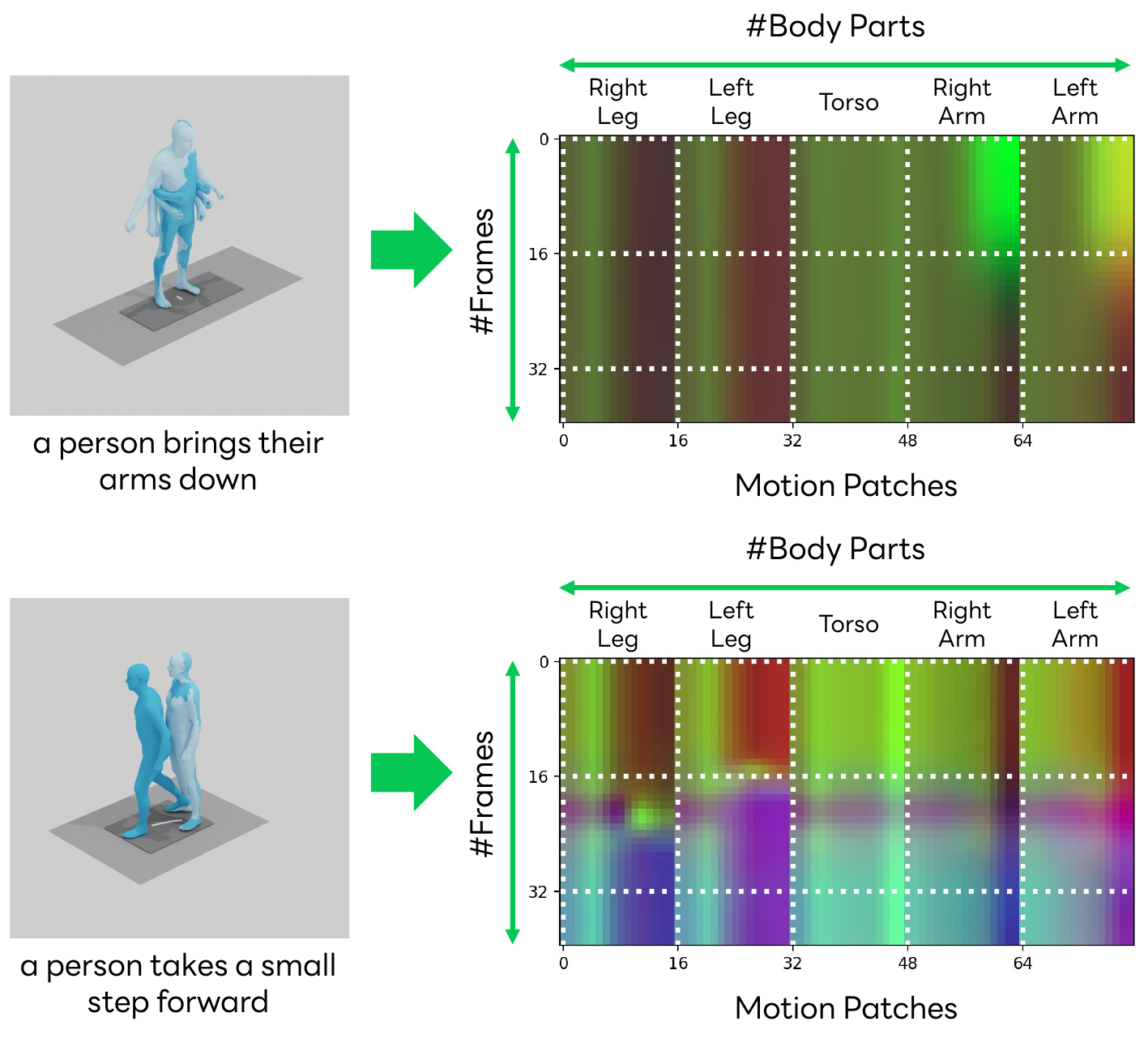}
    \vspace{-10pt}
    \caption{Visualization of motion patches by regarding the joint coordinates as RGB pixels. We show the rendered motions and their text label on the left and the processed motion patches on the right. We can observe different motions reflected in different motion patches.}
    \vspace{-10pt}
    \label{fig:patch_visualization}
\end{figure}

We provide a visualization of motion patches by depicting them as RGB images in~\Fref{fig:patch_visualization}. We interpret joint coordinates as RGB pixels to create a visual representation of each motion patch. The figure displays the rendered motions and their corresponding text labels on the left, and the processed motion patches on the right. We can observe different motions resulting in distinct motion patches, demonstrating the capacity of our method to capture unique characteristics of each motion in the form of motion patches.

\subsection{Motion Encoder}
\label{sec:motion_encoder}
For image data, many well-established architectures and pre-trained models can be used for CLIP. Meanwhile, for motion data, there is no standard architecture and no large-scale pre-trained model. Existing methods~\cite{tevet2022motionclip, petrovich2023tmr} use their original motion encoders and train them from scratch to extract the motion representations. However, with the novel motion patches proposed in~\Sref{sec:motion_patches}, we are able to encode motions by extending ViT for 2D images to 3D motion data to overcome the limited scale of motion data.

ViT first extracts non-overlapping image patches from the image data. Then, a projection head is used to project these patches onto 1D tokens. The transformer architecture is used to model the interaction between each image patch to obtain the final representation. To apply ViT to motion sequences, we first transfer the motion sequences into motion patches and then regard these motion patches as image patches. In this paper, we adopt the ViT-B/16 with 12 layers and the patch size 16 pre-trained on ImageNet-21k~\cite{russakovsky2015imagenet} as our motion encoder. Hence, we set $N=16$ to obtain the motion patches with size $16 \times 16$. We have additionally included an investigation of the choices related to the ViT backbone and patch sizes in the supplementary material.

Following ViT and CLIP, the [\texttt{class}] token is added to the inputs and we resize the position embedding to match the number of patches. The output from the [\texttt{class}] token is projected onto a multi-modal embedding space as the motion representation.

\subsection{Text Encoder}
\label{sec:text_encoder}
In the context of text encoding, it is crucial to extract features related to motion. Following TMR~\cite{petrovich2023tmr}, we adopt DistilBERT~\cite{sanh2019distilbert} for this purpose, utilizing a pre-trained model with a projection head. The output from the [\texttt{class}] token is used as the text representation. An alternative is utilizing the text encoder of CLIP~\cite{radford2021learning}, commonly used in motion generation methods~\cite{tevet2022human, zhang2022motiondiffuse}. However, vision-language models, including CLIP, face challenges in distinguishing between entities and verbs~\cite{hendricks2021probing, yuksekgonul2022and, park2022exposing}. Despite exploring this option, our experiments showed that DistilBERT outperformed CLIP, with detailed comparisons available in the supplementary material.

\subsection{Training Strategy}
\label{sec:training}
Given a batch of $B$ (motion sequence, text) pairs, the model needs to generate and optimize $B \times B$ similarities. We use a symmetric cross-entropy loss over these similarity scores to train the parameters of the model as follows:
\begin{align}
    \mathcal{L}_{m2t} &= -\frac{1}{B} \sum_i^B{\log \frac{\exp(s(m_i, t_i) /\tau)}{\sum_{j=1}^B{\exp(s(m_i, t_j)/\tau)}}}, \\
    \mathcal{L}_{t2m} &= -\frac{1}{B} \sum_i^B{\log \frac{\exp(s(m_i, t_i)/\tau)}{\sum_{j=1}^B{\exp(s(m_j, t_i)/\tau})}}, \\
    \mathcal{L} &= \mathcal{L}_{m2t} + \mathcal{L}_{t2m},
\end{align}
where $\tau$ is the temperature parameter. The loss $\mathcal{L}$ is the sum of motion-to-text loss $\mathcal{L}_{m2t}$ and text-to-motion loss $\mathcal{L}_{t2m}$. 

\section{Experiments}
\label{sec:experiment}

\subsection{Datasets}
We utilize two standard datasets in our experiments: the HumanML3D dataset~\cite{guo2022generating} and the KIT Motion-Language dataset~\cite{plappert2016kit}.

\textbf{HumanML3D Dataset:} The HumanML3D dataset enriches the AMASS~\cite{mahmood2019amass} and HumanAct12~\cite{guo2020action2motion} motion capture collections with natural language labels describing the motions. We follow the same motion pre-processing method proposed in~\cite{guo2022generating}. Furthermore, the dataset is augmented through the mirroring of both left and right motions, along with their corresponding textual descriptions. Subsequently, following the official dataset split, we acquire a total of 23,384, 1,460, and 4,380 motions for the training, validation, and test sets, respectively. On average, each motion receives 3.0 distinct textual annotations. During the training phase, we randomly choose one annotation as the matching text, while for testing, we only use the first one.

\textbf{KIT Motion-Language Dataset (KIT-ML):} The KIT-ML dataset, which primarily focuses on locomotion, is also derived from motion capture data. To prepare the motion data for analysis, we apply the identical pre-processing procedure as employed in the HumanML3D dataset. The dataset is partitioned into training, validation, and test sets, consisting of 4,888, 300, and 830 motions, respectively. On average, each motion is annotated 2.1 times.

\begin{table*}
    \centering
    \setlength{\tabcolsep}{4pt}
    \resizebox{0.99\linewidth}{!}{
    \begin{tabular}{ll|cccccc|cccccc}
        \toprule
         \multirow{2}{*}{\textbf{Protocol}} & \multirow{2}{*}{\textbf{Methods}} & \multicolumn{6}{c|}{Text-motion retrieval} & \multicolumn{6}{c}{Motion-text retrieval} \\
         & & \small{R@1 $\uparrow$} & \small{R@2 $\uparrow$} & \small{R@3 $\uparrow$} &  \small{R@5 $\uparrow$} & \small{R@10 $\uparrow$} & \small{MedR $\downarrow$} & \small{R@1 $\uparrow$} & \small{R@2 $\uparrow$} & \small{R@3 $\uparrow$} & \small{R@5 $\uparrow$} & \small{R@10 $\uparrow$} & \small{MedR $\downarrow$} \\
\hline\hline
\multirow{4}{*}{All} & TEMOS$^\dagger$~\cite{petrovich2022temos} &  2.12 &  4.09 &  5.87 &  8.26 & 13.52 & 173.0 &  3.86 &  4.54 &  6.94 &  9.38 & 14.00 & 183.25 \\
 & T2M$^\dagger$~\cite{guo2022generating} &  1.80 &  3.42 &  4.79 &  7.12 & 12.47 & 81.00 &  2.92 &  3.74 &  6.00 &  8.36 & 12.95 & 81.50 \\
 & TMR~\cite{petrovich2023tmr} &  8.92  & 12.04 & 16.33 & 22.06 & 33.37 & 25.00 & 9.44  & 11.84 & 16.90 & 22.92 & 32.21 & 26.00 \\
 \rowcolor[gray]{0.90} \cellcolor{white} & Ours (scratch)& 8.46 & 12.76 & 16.22 & 23.56 & 35.27 & 23.00 & 9.63 & 11.78 & 16.58 & 22.87 & 33.57 & 25.00\\
 \rowcolor[gray]{0.90} \cellcolor{white} &  Ours &  \textbf{10.80} & \textbf{14.98} & \textbf{20.00} & \textbf{26.72} & \textbf{38.02} & \textbf{19.00} & \textbf{11.25} & \textbf{13.86} & \textbf{19.98} & \textbf{26.86} & \textbf{37.40} & \textbf{20.50}
 \\
\midrule
\multirow{4}{*}{Small batches} & TEMOS$^\dagger$~\cite{petrovich2022temos} & 40.49 & 53.52 & 61.14 & 70.96 & 84.15 &  2.33 & 39.96 & 53.49 & 61.79 & 72.40 & 85.89 &  2.33 \\
& T2M$^\dagger$~\cite{guo2022generating} & 52.48 & 71.05 & 80.65 & 89.66 & 96.58 &  1.39 & 52.00 & 71.21 & 81.11 & 89.87 & 96.78 &  1.38 \\
& TMR~\cite{petrovich2023tmr} & 67.45 & 80.98 & 86.22 & 91.56 & 95.46 & 1.03  & 68.59 & 81.73 & 86.75 & 91.10  & 95.39 & 1.02  \\
 \rowcolor[gray]{0.90} \cellcolor{white} & Ours (scratch) & 67.61 & 82.40 & 86.79 & 91.75 & 95.97 & 1.01 & 67.11 & 80.04 & 85.86 & 91.86 & 95.98 & 1.00 \\
\rowcolor[gray]{0.90} \cellcolor{white} & Ours & \textbf{71.61} & \textbf{85.81} & \textbf{90.02} & \textbf{94.35} & \textbf{97.69} & \textbf{1.00} & \textbf{72.11} & \textbf{85.26} & \textbf{90.21} & \textbf{94.44} & \textbf{97.76} & \textbf{1.00}
 \\
\bottomrule        
\end{tabular}
}
\vspace{-5pt}
\caption{Results of text-to-motion and motion-to-text retrieval benchmark on HumanML3D. The results of methods marked with $\dagger$ are sourced from TMR~\cite{petrovich2023tmr}. Ours (scratch) denotes the proposed method trained from scratch without using pre-trained ViT weights.
}
\vspace{-5pt}
\label{tab:benchmark_h3d}
\end{table*}

\begin{table*}
    \centering
    \setlength{\tabcolsep}{4pt}
    \resizebox{0.99\linewidth}{!}{
    \begin{tabular}{ll|cccccc|cccccc}
        \toprule
         \multirow{2}{*}{\textbf{Protocol}} & \multirow{2}{*}{\textbf{Methods}} & \multicolumn{6}{c|}{Text-motion retrieval} & \multicolumn{6}{c}{Motion-text retrieval} \\
         & & \small{R@1 $\uparrow$} & \small{R@2 $\uparrow$} & \small{R@3 $\uparrow$} & \small{R@5 $\uparrow$} & \small{R@10 $\uparrow$} & \small{MedR $\downarrow$} & \small{R@1 $\uparrow$} & \small{R@2 $\uparrow$} & \small{R@3 $\uparrow$} & \small{R@5 $\uparrow$} & \small{R@10 $\uparrow$} & \small{MedR $\downarrow$} \\
\hline\hline
\multirow{4}{*}{All} & TEMOS$^\dagger$~\cite{petrovich2022temos} &  7.11 & 13.25 & 17.59 & 24.10 & 35.66 & 24.00 & 11.69 & 15.30 & 20.12 & 26.63 & 36.39 & 26.50 \\
 & T2M$^\dagger$~\cite{guo2022generating} &  3.37 &  6.99 & 10.84 & 16.87 & 27.71 & 28.00 &  4.94 &  6.51 & 10.72 & 16.14 & 25.30 & 28.50 \\
 & TMR~\cite{petrovich2023tmr} &  10.05 & 13.87 & 20.74 & 30.03 & 44.66 & 14.00 & 11.83 & 13.74 & 22.14 & 29.39 & 38.55 & 16.00 \\
 \rowcolor[gray]{0.90} \cellcolor{white} & Ours (scratch) & 10.41 & 17.01 & 24.43 & 31.55 & 46.50 & 13.00 & 12.13 & 14.83 & 22.54 & 30.83 & 40.12 & 15.00\\
 \rowcolor[gray]{0.90} \cellcolor{white} & Ours & \textbf{14.02} & \textbf{21.08} & \textbf{28.91} & \textbf{34.10} & \textbf{50.00} & \textbf{10.50} & \textbf{13.61} & \textbf{17.26} & \textbf{27.54} & \textbf{33.33} & \textbf{44.77} & \textbf{13.00 }\\
\midrule
\multirow{4}{*}{Small batches} & TEMOS$^\dagger$~\cite{petrovich2022temos} & 43.88 & 58.25 & 67.00 & 74.00 & 84.75 &  2.06 & 41.88 & 55.88 & 65.62 & 75.25 & 85.75 &  2.25 \\
& T2M$^\dagger$~\cite{guo2022generating} & 42.25 & 62.62 & 75.12 & 87.50 & 96.12 &  1.88 & 39.75 & 62.75 & 73.62 & 86.88 & 95.88 &  1.95 \\
& TMR~\cite{petrovich2023tmr} & 50.00 & 69.14 & 78.02 & 87.97 & 94.87 & 1.50  & 51.21 & 69.53 & 78.64 & 89.00 & 95.31 & 1.50  \\
 \rowcolor[gray]{0.90} \cellcolor{white} & Ours (scratch) & 51.13 & 70.15 & 78.96 & 88.06&95.74&1.43&53.12&70.31&79.40&89.34&95.59&1.33\\
\rowcolor[gray]{0.90} \cellcolor{white} & Ours & \textbf{53.55} & \textbf{71.30} & \textbf{79.82} & \textbf{88.92} & \textbf{96.29} & \textbf{1.36} & \textbf{54.54} & \textbf{72.15} & \textbf{79.68} & \textbf{89.35} & \textbf{96.11} & \textbf{1.31}
\\
\bottomrule        
\end{tabular}
}
\vspace{-5pt}
\caption{Results of text-to-motion and motion-to-text retrieval benchmark on KIT-ML.}
\vspace{-5pt}
\label{tab:benchmark_kitml}
\end{table*}

\subsection{Evaluation Protocol}
\label{sec:evaluation_protocol}
To evaluate the performance of the motion-language model, we adopt the retrieval task between the motion sequence and the text description. Following~\cite{petrovich2023tmr}, our evaluation of retrieval performance employs standard metrics, specifically Recall at various ranks (R@1, R@2, etc.), for both text-to-motion and motion-to-text tasks. Recall at rank $k$ indicates the percentage of instances where the correct label appears within the top $k$ results \footnote{Due to the existence of mirroring augmented samples in the test data, some samples have two correct answers in the gallery, \ie, the original motion and its mirrored counterpart may share the same text description. We accounted for this factor, whereas TMR~\cite{petrovich2023tmr} metrics overlooked it.}, with higher values indicating better performance. Additionally, we calculate the median rank (MedR), where a lower value shows better performance. It is important to note that the evaluation of retrieval performance is conducted using an unseen gallery of real motions, specifically the test set. 

We used several evaluation protocols to calculate Recall, primarily altering the composition of the gallery set:

\textbf{All:} In this protocol, the entire test set is used without any modifications. However, repetitive texts across motions or minor textual differences (\eg, ``person'' vs. ``human'', ``walk'' vs. ``walking'') will affect the results. We use this protocol as the default protocol in this paper.

\textbf{Small Batches:} This protocol is designed by Guo et al.~\cite{guo2022generating}. It involves randomly selecting batches of 32 motion-text pairs and then reporting the average performance. While this approach introduces randomness, it serves as a benchmark for comparison. It is worth noting that a gallery size of 32 is relatively manageable compared to the other protocols, making it a less challenging scenario.

\subsection{Implementation Details}
In our experiments, we employ the Adam optimizer~\cite{kingma2014adam} with a learning rate of $10^{-5}$ for the text encoder,  $10^{-4}$ for the motion encoder, and $10^{-3}$ for the projection head. A batch size of 256 is used during the training. The latent dimension of the embeddings after the projection is set to 256. We set the temperature parameter to 0.07 following CLIP. The number of frames in each motion sequence is limited to 224, following the existing methods~\cite{tevet2022motionclip, petrovich2023tmr}, which means $14\times5=70$ motion patches are used as the input of ViT.

\subsection{Results}
In our evaluation of text-to-motion and motion-to-text retrieval benchmark across HumanML3D (Table~\ref{tab:benchmark_h3d}) and KIT-ML (Table~\ref{tab:benchmark_kitml}) datasets, encompassing all evaluation protocols, we provide comparisons against prior works, specifically TEMOS~\cite{petrovich2022temos}, T2M~\cite{guo2022generating}, TMR~\cite{petrovich2023tmr} and the proposed method trained from scratch without using pre-trained ViT weights. The experimental results of TEMOS~\cite{petrovich2022temos} and T2M~\cite{guo2022generating} are sourced from the TMR~\cite{petrovich2023tmr} paper. Meanwhile, we re-evaluate the official models of TMR~\cite{petrovich2023tmr} using our evaluation code to ensure a fair comparison.

It is important to note that TEMOS~\cite{petrovich2022temos} is not explicitly designed for retrieval tasks. The cross-modal embedding space of TEMOS~\cite{petrovich2022temos} is primarily trained with positive pairs. In contrast, T2M~\cite{guo2022generating} applied their method to retrieval by employing contrastive learning, which includes negative pairs as well. TMR~\cite{petrovich2023tmr} is the state-of-the-art method for text-to-motion retrieval, which extends TEMOS by incorporating a contrastive loss~\cite{oord2018representation} between the motion features and the text features in the latent space.

Remarkably, our model consistently outperforms prior work across all evaluation sets in various degrees of difficulty. This indicates that our model can capture the nuanced nature of motion descriptions. The substantial performance enhancements we achieve over the state-of-the-art can be attributed to several factors: (1) the design of motion patches to capture the temporal-spatial motion representation and (2) the utilization of ViT and transferring its pre-trained weights to the motion domain. In the subsequent sections, we conduct controlled experiments to analyze the impact of these components on our results.

\subsection{Ablation Studies}
In this section, we explore various settings to better understand the factors influencing the performance of our model.

\textbf{Pre-trained ViT:} We conduct experiments to compare the performance of our model when utilizing pre-trained ViT representations against a setting where ViT pre-training is not employed. 

\textbf{Motion Representation:} Another aspect we investigate is the use of motion representations. We investigate different scenarios where we either employ the proposed motion patches as the input for ViT or directly feed the raw motion sequences into the Transformer component, along with positional encodings. 

The results of HumanML3D and KIT-ML are shown in~\Tref{tab:ablation_study}. It is noticeable that when the motion patches are used, training the model from pre-trained weights leads to much better results than training from scratch. Compared with using motion patches as input, using motion sequences without preprocessing leads to worse performance. This analysis shows the impact of ViT pre-training on the capabilities of our model and the advantages that our motion patches bring to retrieval tasks.
\begin{table}
    \centering
    \setlength{\tabcolsep}{2pt}
    \resizebox{0.99\linewidth}{!}{
    \begin{tabular}{cc|cccc|cccc}
        \multicolumn{10}{c}{Dataset: HumanML3D} \\
        \toprule
            \textbf{Pre-trained} & \textbf{Motion} & \multicolumn{4}{c|}{Text-motion retrieval} & \multicolumn{4}{c}{Motion-text retrieval} \\
            \textbf{ViT} & \textbf{Patches} & \small{R@1 $\uparrow$} & \small{R@5 $\uparrow$} & \small{R@10 $\uparrow$}  & \small{MedR $\downarrow$} & \small{R@1 $\uparrow$} & \small{R@5 $\uparrow$} & \small{R@10 $\uparrow$} & \small{MedR $\downarrow$} \\
        \hline\hline
        \cmark & \cmark & \textbf{10.80} & \textbf{26.72} & \textbf{38.02} & \textbf{19.00} & \textbf{11.25} & \textbf{26.86} & \textbf{37.40} & \textbf{20.50} \\
        \xmark & \cmark &  8.46 & 23.56 & 35.37 & 23.00 & 9.63 & 22.87 & 33.57 & 25.00 \\
        \cmark & \xmark & 8.36 & 22.84 & 33.62 & 24.00 & 8.81 & 21.67 & 31.35 & 29.00 \\
        \xmark & \xmark & 8.58 & 21.54 & 32.87 & 25.00 & 8.46 & 21.96 & 30.79 & 30.00 \\
        \bottomrule        
    \multicolumn{10}{c}{} \\
    \multicolumn{10}{c}{Dataset: KIT-ML} \\
        \toprule
            \textbf{Pre-trained} & \textbf{Motion} & \multicolumn{4}{c|}{Text-motion retrieval} & \multicolumn{4}{c}{Motion-text retrieval} \\
            \textbf{ViT} & \textbf{Patches} & \small{R@1 $\uparrow$} & \small{R@5 $\uparrow$} & \small{R@10 $\uparrow$}  & \small{MedR $\downarrow$} & \small{R@1 $\uparrow$} & \small{R@5 $\uparrow$} & \small{R@10 $\uparrow$} & \small{MedR $\downarrow$} \\
        \hline\hline
        \cmark & \cmark & \textbf{14.02} & \textbf{34.10} & \textbf{50.00} & \textbf{10.50} & \textbf{13.61} & \textbf{33.33} & \textbf{44.77} & \textbf{13.00 }\\
        \xmark & \cmark &  10.41 & 31.55 & 46.50 & 13.00 & 12.13 & 30.83 & 40.12 & 15.00\\
        \cmark & \xmark &  9.54 & 31.72 & 45.86 & 14.00 & 11.01 & 27.31 & 37.41 & 18.00 \\
        \xmark & \xmark & 9.87 & 30.37 & 44.21 & 16.00 & 10.14 & 26.86 & 36.43 & 21.00 \\
        \bottomrule        
    \end{tabular}   
    }
    \vspace{-2pt}
    \caption{Results of ablation studies. We experiment with different settings (1) with/without the pre-trained ViT and (2) whether to use motion patches as the representation of the motion. 
    }
    \vspace{-2pt}
    \label{tab:ablation_study}
\end{table}

\subsection{Qualitative results}

\begin{figure}[t]
    \centering
    \includegraphics[width=1\linewidth]{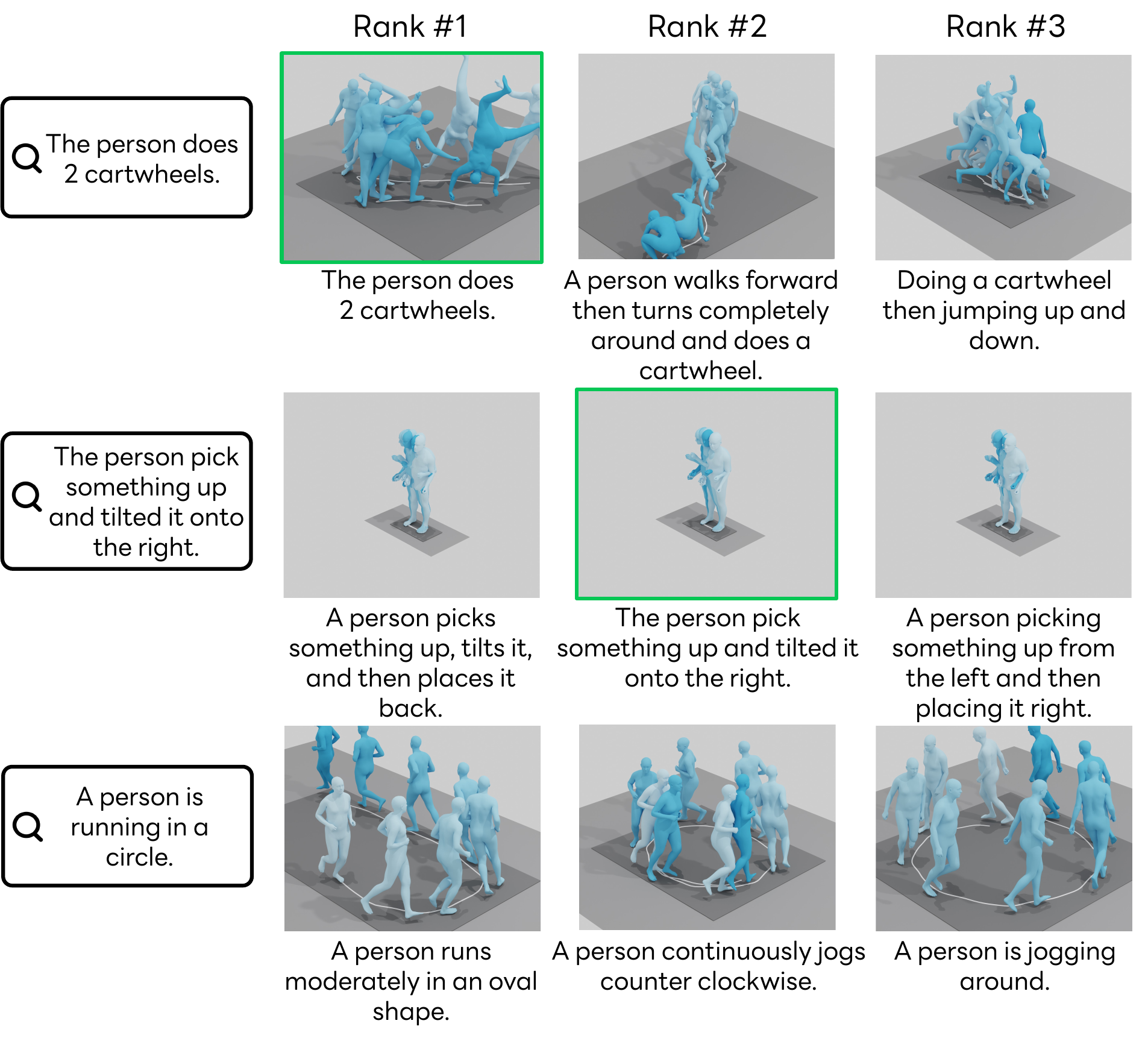}
    \vspace{-5pt}
    \caption{Qualitative results of text-to-motion retrieval. For each query, we show the retrieved motions ranked by text-motion similarity and their accompanying ground-truth text labels. Note that these descriptions are not used in the retrieval process. All motions in the gallery are from the test set and were unseen during training. For the first two examples, the text queries are sampled from the data. For the last example, we query with a free-form text.}
    \vspace{-5pt}
    \label{fig:visualization}
\end{figure}

In~\Fref{fig:visualization}, we present the qualitative results of text-to-motion retrieval on the entire test set of HumanML3D. Each query text is displayed on the left, and on the right, we showcase the top-3 retrieved motions along with their corresponding ground-truth text labels. The gallery of motions for retrieval remains unseen during training.

In the first two examples, we successfully retrieve the ground-truth motion in the top-2 results. Note that in the second example, the differences between the motions are very small and they all present motions similar to the text query. For the free-form prompt in the last example, where the exact text is not present in the gallery, our method also succeeds in retrieving correct motions.

\section{Applications}
\label{sec:application}

\subsection{Cross-skeleton Recognition}
\label{sec:cross_skeleton}
One advantage of the proposed motion patches is that the motion sequences from different skeleton structures can be transferred into a unified representation. For example, motion in HumanML3D~\cite{guo2022generating} follows the skeleton structure of SMPL~\cite{loper2015smpl} with 22 joints as shown in~\Fref{fig:motion_patch}, while poses have 21 joints in KIT-ML, and the skeleton structure of KIT-ML is different from SMPL (detailed in the supplementary). Existing methods cannot deal with these two datasets simultaneously because the dimension of the input vector varies according to the dimension of the pose vector. However, our method is able to convert the motion sequence to 16$\times$16 motion patches according to the kinematic chain of each body part, which means the motion features learned on a dataset can be transferred to another dataset even when the skeleton structure is different.

To evaluate the performance of the proposed method for cross-skeleton recognition, we prepare two scenarios. The first one is a zero-shot setting, where we directly apply the motion-language model trained on the HumanML3D dataset to the text-to-motion retrieval task of the KIT-ML dataset. The other one is a transfer learning setting, where we further fine-tune the HumanML3D model with the KIT-ML dataset, because the scale of the KIT-ML dataset is smaller than that of the HumanML3D dataset. 

The results are shown in~\Tref{tab:cross_skeleton}, which compares the zero-shot setting and the transfer learning setting with other existing methods and the proposed method trained on KIT-ML. The performance of zero-shot prediction with the HumanML3D model is lower than the models trained on KIT-ML, because the language domains (\ie, KIT-ML is more focused on locomotion descriptions) and the skeleton structures of these two datasets are different. However, it still achieves acceptable performance compared to TEMOS~\cite{petrovich2022temos} and T2M~\cite{guo2022generating}, especially on the task of text-motion retrieval. It is noticeable that the transfer learning method obtained the best results, which shows that training the model with HumanML3D is helpful in recognizing the motion sequences of KIT-ML. This highlights the potential of our approach to improve the performance on small-scale datasets by pre-training the model on large-scale datasets.

\begin{table}
    \centering
    \setlength{\tabcolsep}{2pt}
    \resizebox{0.99\linewidth}{!}{
    \begin{tabular}{cc|cccc|cccc}
    \multicolumn{10}{c}{Dataset: KIT-ML} \\
        \toprule
            \multirow{2}{*}{\textbf{Method}} & \textbf{Training} & \multicolumn{4}{c|}{Text-motion retrieval} & \multicolumn{4}{c}{Motion-text retrieval} \\
            & \textbf{Dataset} & \small{R@1 $\uparrow$} & \small{R@5 $\uparrow$} & \small{R@10 $\uparrow$}  & \small{MedR $\downarrow$} & \small{R@1 $\uparrow$} & \small{R@5 $\uparrow$} & \small{R@10 $\uparrow$} & \small{MedR $\downarrow$} \\
        \hline\hline
        TEMOS$^\dagger$~\cite{petrovich2022temos} & KIT-ML & 7.11 & 24.10 & 35.66 & 24.00 & 11.69 & 26.63 & 36.39 & 26.50 \\
        T2M$^\dagger$~\cite{guo2022generating} & KIT-ML & 3.37 & 16.87 & 27.71 & 28.00 & 4.94 & 16.14 & 25.30 & 28.50 \\
        TMR~\cite{petrovich2023tmr} & KIT-ML & 10.05 & 30.03 & 44.66 & 14.00 & 11.83 & 29.39 & 38.55 & 16.00 \\
        Ours & KIT-ML & 14.02 & 34.10 & 50.00 & 10.50 & 13.61 & 33.33 & 44.77 & 13.00\\ \midrule
        Ours & HumanML3D & 7.35 & 20.98 & 34.33 & 23.50 & 7.90 & 17.71 & 25.88 & 31.00 \\
        \rowcolor[gray]{0.90} Ours & Transferred & \textbf{15.28} & \textbf{38.71} & \textbf{52.65} & \textbf{9.00} & \textbf{16.51} & \textbf{35.78} & \textbf{47.47} & \textbf{11.00} \\
        \bottomrule        
    \end{tabular}   
    }
    \vspace{-2pt}
    \caption{Results of cross-skeleton recognition. We evaluate the text-to-motion and motion-to-text retrieval on the KIT-ML dataset with the HumanML3D model and the transferred model. The transfer learning method achieves better performance than the method of training only on KIT-ML.
    }
    \vspace{-2pt}
    \label{tab:cross_skeleton}
\end{table}

\subsection{Zero-shot Motion Classification}
Furthermore, we demonstrate the effectiveness of the semantically structured latent spaces generated by our motion-language model via action recognition. We follow the BABEL 60-classes benchmark~\cite{punnakkal2021babel}, containing 10,892 sequences, and 20\% of them are used as test sets. We preprocessed the motion sequences with the same procedure as HumanML3D~\cite{guo2022generating}. Because this is a zero-shot classification setting, we did not train the model with BABEL and only applied the model trained on HumanML3D to the test data. For the text prompts, the action names in BABEL are used as ``A person \{action\}''. We calculate the cosine distance between a given motion and all 60 text prompts.

In~\Tref{tab:zero-shot}, we show a comparison of the Top-1 and Top-5 accuracy achieved by our zero-shot classifier with that of the 2s-AGCN classifier~\cite{punnakkal2021babel} and MotionCLIP~\cite{tevet2022motionclip}. As evident from the results, our framework performs comparably to the state-of-the-art supervised methods, despite the fact that our method was not initially designed for action recognition tasks nor trained on the action label set of BABEL. This highlights the versatility and adaptability of our approach across various applications.

\begin{table}[t]
\begin{center}
\tabcolsep = 1.0mm
\resizebox{0.99\linewidth}{!}{
\begin{tabular}{c|ccc|cc}
\toprule
\textbf{Method} & \begin{tabular}[c]{@{}c@{}}\textbf{Training} \\ \textbf{Dataset} \end{tabular} & \textbf{Zero-shot} & \textbf{Modality} & \textbf{Top-1 Acc.} & \textbf{Top-5 Acc.} \\\hline\hline
2s-AGCN~\cite{shi2019two} & BABEL & \xmark & M & 41.14 & \textbf{73.18} \\
MotionCLIP~\cite{tevet2022motionclip} & BABEL & \xmark & M+L & 40.90 &  57.71 \\
TMR~\cite{petrovich2023tmr} & HumanML3D & \cmark & M+L & 30.13 &  41.52
\\
\rowcolor[gray]{0.90} Ours & HumanML3D & \cmark & M+L & \textbf{41.33} &  68.97 \\
\bottomrule
\end{tabular}
}
\end{center}
\vspace{-12pt}
\caption{Results of zero-shot motion classification. Modality with motion only and motion language are denoted as M and M+L, respectively. When applying our proposed method for zero-shot classification, we achieve performance results that are closely aligned with those of the 2s-AGCN classifier trained with supervision on the BABEL-60 benchmark.}

\label{tab:zero-shot}
\end{table}

\begin{table}
    \centering
    \setlength{\tabcolsep}{2pt}
    \resizebox{0.99\linewidth}{!}{
    \begin{tabular}{c|cccc|cccc}
        \toprule
            \multirow{2}{*}{\textbf{Method}} & \multicolumn{4}{c|}{Text-motion retrieval} & \multicolumn{4}{c}{Motion-text retrieval} \\
            & \small{R@1 $\uparrow$} & \small{R@5 $\uparrow$} & \small{R@10 $\uparrow$}  & \small{MedR $\downarrow$} & \small{R@1 $\uparrow$} & \small{R@5 $\uparrow$} & \small{R@10 $\uparrow$} & \small{MedR $\downarrow$} \\
        \hline\hline
        TMR~\cite{petrovich2023tmr} & 5.38 & 15.64 & 24.40 & 34.00 & 5.13 & 15.26 & 25.65 & 33.00 \\
        \rowcolor[gray]{0.90} Ours & \textbf{9.51} & \textbf{21.27} & \textbf{32.41} & \textbf{27.00} & \textbf{8.26} & \textbf{22.65} & \textbf{32.66} & \textbf{24.00} \\
        \bottomrule          
    \end{tabular}   
    }
    \vspace{-2pt}
    \caption{Results of human interaction recognition. For TMR~\cite{petrovich2023tmr} and our method, we concatenate the motion features of each person and get the multi-person motion feature through a projection head of the concatenated feature.}
    
    \label{tab:interaction}
\end{table}

\subsection{Human Interaction Recognition}
Besides the recognition of single-person motion, the proposed method can also be applied to multi-person motion recognition. We conduct the experiments using InterHuman~\cite{liang2023intergen}, a motion-language dataset that comprises a diverse set of 3D motions involving interactions between two individuals. The dataset is split into 6,222 sequences for training and 1,557 sequences for testing. We processed the motion sequences of each individual with the same procedure as HumanML3D~\cite{guo2022generating}. To obtain the features of the interactions, we apply a shared motion encoder to each person and simply concatenate the motion features before the projection head. We evaluate the performance of the proposed method via text-to-motion retrieval. The results are shown in~\Tref{tab:interaction} and our method outperforms TMR~\cite{petrovich2023tmr}.
\section{Limitations}
\label{sec:limitation}
In this paper, we primarily evaluate our method on motion recognition, focusing on text-to-motion retrieval. Future work includes applying our method to text-to-motion generation. Despite leveraging pre-trained vision models to handle small-scale motion datasets, the generalization performance of our method may be limited due to the comparatively smaller size of motion-text data versus image-text data. However, our proposed motion patch, a skeleton-robust representation, aids in constructing large-scale motion datasets from diverse motion capture systems.
\section{Conclusion}
\label{sec:conclusion}
In this paper, we introduced a novel unified motion representation called ``motion patches'' and applied the ViT architecture with its pre-trained weights to build motion-language models. Our approach effectively addresses challenges related to limited data scales in 3D human motion data and diverse skeleton structures, characterized by complex spatial-temporal dependencies. As a result, we have made significant advancements in motion recognition, including text-to-motion retrieval and other applications.
\clearpage
{
    \small
    \bibliographystyle{ieeenat_fullname}
    \bibliography{main}
}

\appendix
\clearpage
\setcounter{page}{1}
\maketitlesupplementary

\section{Motion Patches}
\label{suppl_sec:motion_patches}

\begin{figure}[t]
    \centering
    \includegraphics[width=1\linewidth]{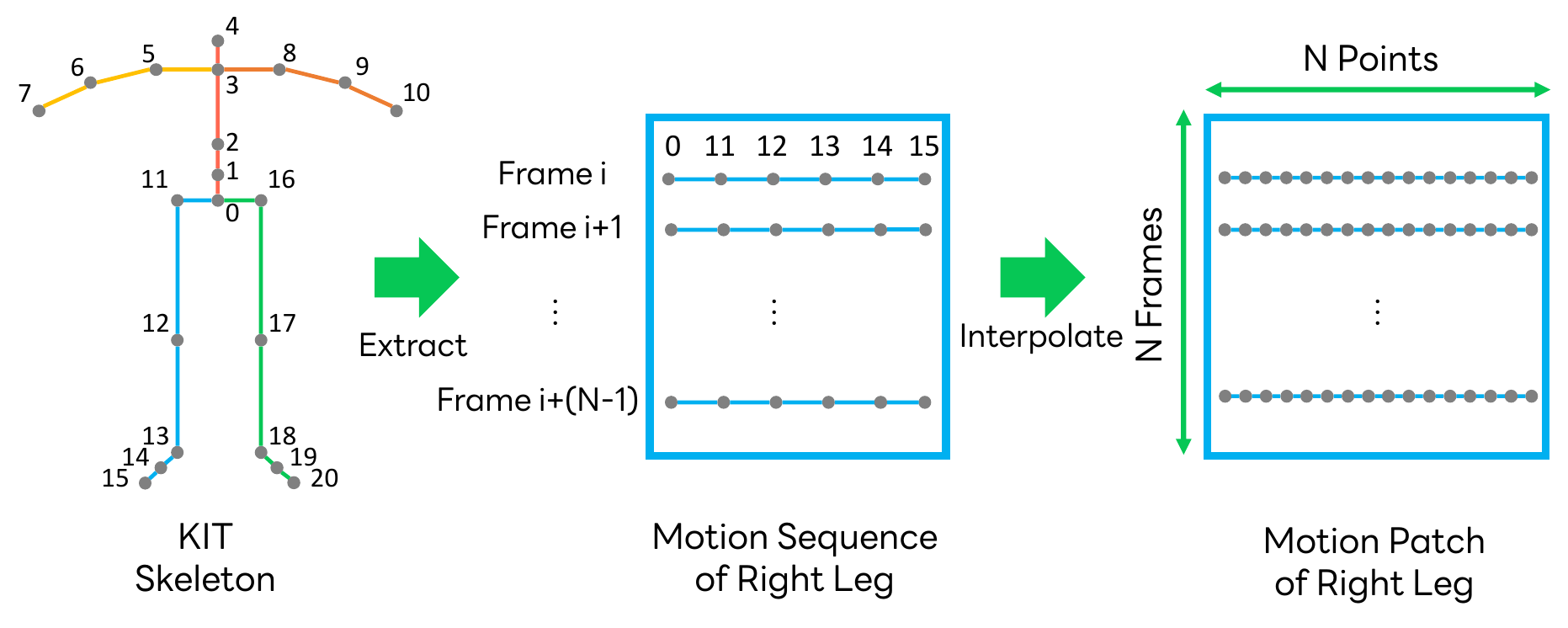}
    \caption{The process of building the motion patches for each motion sequence in KIT-ML. Different body parts are colored in different colors. We show the method to construct the motion patch of the right leg. The same process is applied to other body parts.}
    \label{fig:motion_patch_kit}
\end{figure}

In~\Fref{fig:motion_patch}, we show the process of constructing motion patches for SMPL skeletons in the HumanML3D dataset. For the KIT-ML dataset, the skeleton structure is different but the process is the same as shown in~\Fref{fig:motion_patch_kit}. Because the motion patches use the kinematic chain of the skeleton to extract the spatial-temporal information in motion sequences, our model can be used in cross-skeleton recognition as detailed in~\Sref{sec:cross_skeleton} of the main paper.

\section{Additional Experimental Results}
\subsection{Visualization of Attention Maps}
\begin{figure}[t]
    \centering
    \includegraphics[width=1\linewidth]{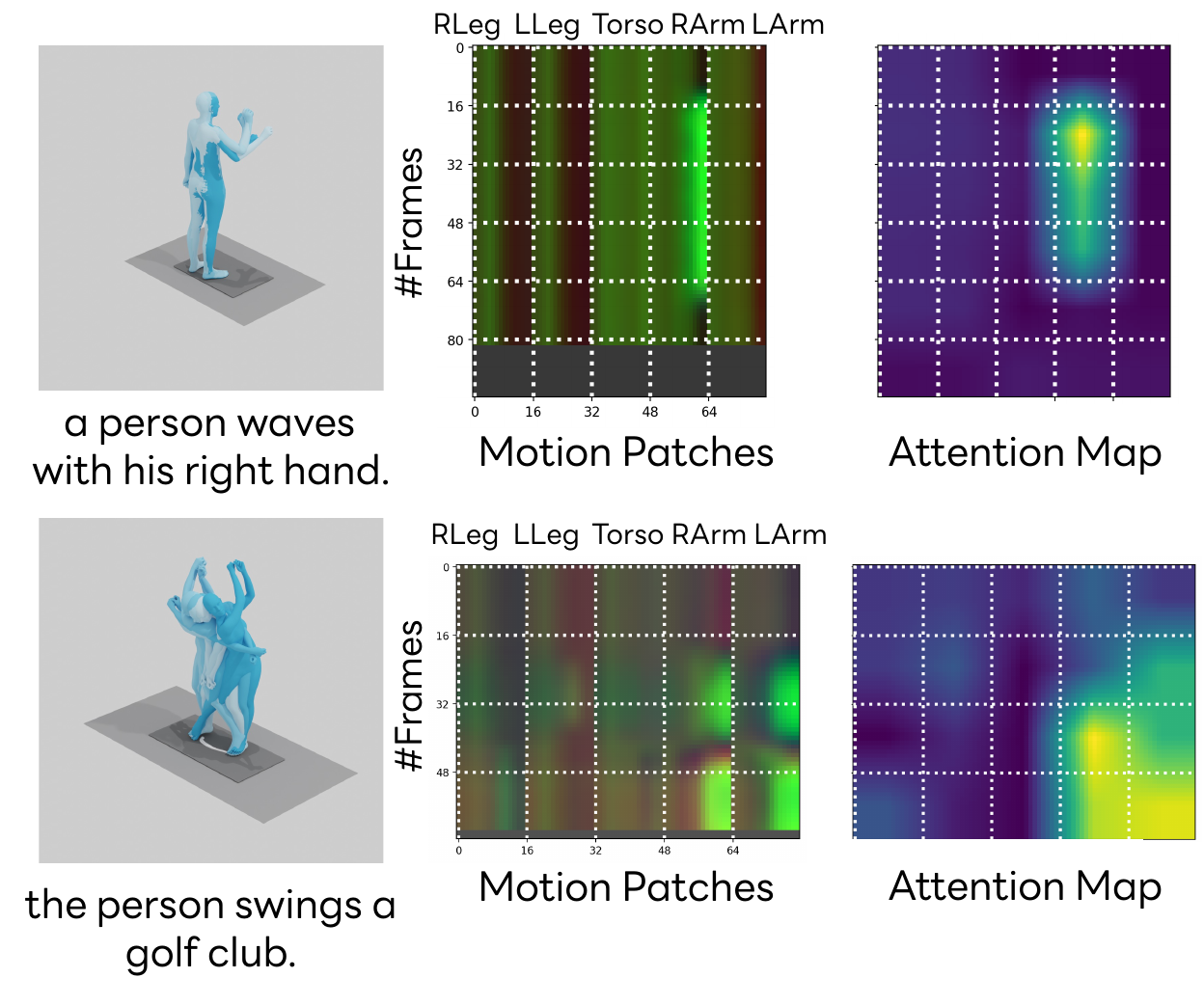}
    \caption{Visualization of attention maps extracted from ViT.}
    \label{fig:attentions_maps}
\end{figure}

In this paper, we find that pre-trained image ViT can help the learning of motion data with the proposed motion patches. As shown in~\Fref{fig:patch_visualization}, the motion patches can be regarded as a kind of spectrogram, where certain patterns related to motions can be observed. Pre-trained ViT helps detect these patterns, which makes transfer learning work. We additionally visualize the attention maps extracted from the ViT trained by our method in~\Fref{fig:attentions_maps}, where the important patterns are activated in the attentions. An analogous approach is audio recognition by rendering the spectrogram of audio as the input into pre-trained image models~\cite{gong2021psla}.

\subsection{ViT Backbones}
We evaluated our method with different ViT backbones. In the main paper, we used ViT-B/16 as the motion encoder. We additionally tried ViT with tiny, small, and large sizes provided in TIMM~\footnote{\url{https://github.com/rwightman/pytorch-image-models}}, and the results are shown in~\Tref{tab:vit_backbone}. We can find that ViT-Tiny and ViT-Small perform a little worse when compared to ViT-base in both datasets. The largest model, ViT-Large, performs well in the HumanML3D dataset, but not well in the KIT-ML Dataset, which may be due to the limited scale of the data. Overall, our proposed method works well on all the ViT backbones. 

\subsection{Motion and Text Encoders}

In the paper, we employed the ViT pre-trained on ImageNet as the motion encoder and the pre-trained DistilBERT~\cite{sanh2019distilbert} as the text encoder. Additionally, we explored an alternative approach by utilizing the image encoder and text encoder of CLIP~\cite{radford2021learning} as the motion and text encoders in our method for comparison. The results are shown in~\Tref{tab:encoder}. We can find that the pre-trained weights affect the performance of the model and the combination of ViT with ImageNet and DistilBERT achieved the best results. When the model of CLIP is used as the motion encoder or the text encoder, we find that the performance drops a little, which shows that CLIP is not effective for capturing motion representations. This might be because CLIP is pre-trained to focus on the semantic features of real-world images, while the motion patches resemble a type of spectrogram with color patterns.

\subsection{Sizes of Motion Patches}
Our investigation explores various motion patch sizes as detailed in~\Tref{tab:patch_size}. In addition to the 16$\times$16 motion patches described in the paper, we have implemented our approach using 8$\times$8 and 32$\times$32 motion patches. Interestingly, both 8$\times$8 and 32$\times$32 patches yielded favorable results. Nevertheless, it is s worth noting that the 16$\times$16 patches consistently delivered the best overall performance.

\begin{table}
    \centering
    \setlength{\tabcolsep}{2pt}
    \resizebox{0.99\linewidth}{!}{
    \begin{tabular}{c|cccc|cccc}
        \multicolumn{9}{c}{Dataset: HumanML3D} \\
        \toprule
            \multirow{2}{*}{\textbf{ViT Size}} & \multicolumn{4}{c|}{Text-motion retrieval} & \multicolumn{4}{c}{Motion-text retrieval} \\
             & \small{R@1 $\uparrow$} & \small{R@5 $\uparrow$} & \small{R@10 $\uparrow$}  & \small{MedR $\downarrow$} & \small{R@1 $\uparrow$} & \small{R@5 $\uparrow$} & \small{R@10 $\uparrow$} & \small{MedR $\downarrow$} \\
        \hline\hline
        Tiny & 9.54 & 23.77 & 36.10 & 20.00 & 10.52 & 24.00 & 33.09 & 24.00 \\
        Small & 9.63 & 24.64 & 36.85 & 19.00 & 10.30 & 24.04 & 33.77 & 23.00 \\
        \rowcolor[gray]{0.90} Base & \textbf{10.80} & 26.72 & 38.02 & \textbf{18.00} & 11.25 & \textbf{26.86} & 37.40 & 19.50 \\
        Large & 10.47 & \textbf{27.29} & \textbf{38.84} & 19.00 & \textbf{11.33} & 26.82 & \textbf{37.42} & \textbf{19.00}\\
        \bottomrule        
    \multicolumn{9}{c}{} \\
    \multicolumn{9}{c}{Dataset: KIT-ML} \\
        \toprule
            \multirow{2}{*}{\textbf{ViT Size}}  & \multicolumn{4}{c|}{Text-motion retrieval} & \multicolumn{4}{c}{Motion-text retrieval} \\
             & \small{R@1 $\uparrow$} & \small{R@5 $\uparrow$} & \small{R@10 $\uparrow$}  & \small{MedR $\downarrow$} & \small{R@1 $\uparrow$} & \small{R@5 $\uparrow$} & \small{R@10 $\uparrow$} & \small{MedR $\downarrow$} \\
        \hline\hline
        Tiny & 11.84 & 33.38 & 48.48 & 11.50 & 12.53 & 28.89 & 40.83 & 16.00 \\
        Small & 12.06 & 33.45 & 49.00 & 11.00 & 13.94 & 28.80 & 39.51 & 17.00 \\
        \rowcolor[gray]{0.90} Base & 14.02 & \textbf{34.10} & \textbf{50.00} & \textbf{10.50} & \textbf{13.61} & \textbf{33.33} & \textbf{44.77} & \textbf{13.00 }\\
        Large & \textbf{14.46} & 32.53 & 42.77 & 15.00 & 13.49 & 28.80 & 38.31 & 18.00 \\
        \bottomrule        
    \end{tabular}   
    }
    \vspace{-5pt}
    \caption{Results of retrieval with different ViT backbones.
    }
    \vspace{-5pt}
    \label{tab:vit_backbone}
\end{table}

\begin{table}
    \centering
    \setlength{\tabcolsep}{2pt}
    \resizebox{0.99\linewidth}{!}{
    \begin{tabular}{cc|cccc|cccc}
        \multicolumn{10}{c}{Dataset: HumanML3D} \\
        \toprule
            \textbf{Motion} & \textbf{Text} & \multicolumn{4}{c|}{Text-motion retrieval} & \multicolumn{4}{c}{Motion-text retrieval} \\
            \textbf{Encoder} & \textbf{Encoder} & \small{R@1 $\uparrow$} & \small{R@5 $\uparrow$} & \small{R@10 $\uparrow$}  & \small{MedR $\downarrow$} & \small{R@1 $\uparrow$} & \small{R@5 $\uparrow$} & \small{R@10 $\uparrow$} & \small{MedR $\downarrow$} \\
        \hline\hline
        \rowcolor[gray]{0.90} \begin{tabular}{c}ViT (ImageNet)\end{tabular}& DistilBERT & \textbf{10.80} & \textbf{26.72} & \textbf{38.02} & \textbf{18.00} & \textbf{11.25} & \textbf{26.86} & \textbf{37.40} & \textbf{19.50} \\
        \begin{tabular}{c}ViT (ImageNet)\end{tabular}& CLIP & 9.66 & 24.12 & 35.47 & 21.00 & 10.37 & 24.50 & 34.35 & 24.00 \\
        \begin{tabular}{c}ViT (CLIP)\end{tabular}& DistilBERT & 9.85 & 24.93 & 36.16 & 21.00 & 10.23 & 24.31 & 34.03 & 23.00 \\
        \begin{tabular}{c}ViT (CLIP)\end{tabular}& CLIP & 6.84 & 18.57 & 29.45 & 32.00 & 7.82 & 19.12 & 27.41 & 35.00 \\
        \bottomrule        
    \multicolumn{10}{c}{} \\
    \multicolumn{10}{c}{Dataset: KIT-ML} \\
        \toprule
            \textbf{Motion} & \textbf{Text}  & \multicolumn{4}{c|}{Text-motion retrieval} & \multicolumn{4}{c}{Motion-text retrieval} \\
            \textbf{Encoder} & \textbf{Encoder} & \small{R@1 $\uparrow$} & \small{R@5 $\uparrow$} & \small{R@10 $\uparrow$}  & \small{MedR $\downarrow$} & \small{R@1 $\uparrow$} & \small{R@5 $\uparrow$} & \small{R@10 $\uparrow$} & \small{MedR $\downarrow$} \\
        \hline\hline
        \rowcolor[gray]{0.90} \begin{tabular}{c}ViT (ImageNet)\end{tabular}& DistilBERT & \textbf{14.02} & \textbf{34.10} & \textbf{50.00} & \textbf{10.50} & \textbf{13.61} & \textbf{33.33} & \textbf{44.77} & \textbf{13.00 }\\
        \begin{tabular}{c}ViT (ImageNet)\end{tabular}& CLIP & 13.01 & 33.29 & 49.76 & 11.00 & 12.66 & 31.45 & 41.45 & 16.00 \\
        \begin{tabular}{c}ViT (CLIP)\end{tabular}& DistilBERT & 10.60 & 32.77 & 45.54 & 13.00 & 12.89 & 26.63 & 37.83 & 18.00 \\
        \begin{tabular}{c}ViT (CLIP)\end{tabular}& CLIP & 10.48 & 26.51 & 36.75 & 24.00 & 11.69 & 23.61 & 30.48 & 36.00 \\
        \bottomrule     
    \end{tabular}   
    }
     \vspace{-5pt}
    \caption{Results of retrieval with different motion and text encoders.
    }
     \vspace{-5pt}
    \label{tab:encoder}
\end{table}

\begin{table}
    \centering
    \setlength{\tabcolsep}{2pt}
    \resizebox{0.99\linewidth}{!}{
    \begin{tabular}{c|cccc|cccc}
        \multicolumn{9}{c}{Dataset: HumanML3D} \\
        \toprule
            \multirow{2}{*}{\textbf{Patch Size}} & \multicolumn{4}{c|}{Text-motion retrieval} & \multicolumn{4}{c}{Motion-text retrieval} \\
             & \small{R@1 $\uparrow$} & \small{R@5 $\uparrow$} & \small{R@10 $\uparrow$}  & \small{MedR $\downarrow$} & \small{R@1 $\uparrow$} & \small{R@5 $\uparrow$} & \small{R@10 $\uparrow$} & \small{MedR $\downarrow$} \\
        \hline\hline
        8$\times$8 & 9.80 & 26.60 & \textbf{38.15} & \textbf{18.00} & \textbf{11.74} & 26.05 & 36.76 & \textbf{19.00} \\
        \rowcolor[gray]{0.90} 16$\times$16 & \textbf{10.80} & \textbf{26.72} & 38.02 & \textbf{18.00} & 11.25 & \textbf{26.86} & \textbf{37.40} & 19.50 \\
        32$\times$32 & 10.13 & 26.22 & 38.00 & 20.00 & 10.90 & 24.88 & 34.82 & 22.00 \\
        \bottomrule        
    \multicolumn{9}{c}{} \\
    \multicolumn{9}{c}{Dataset: KIT-ML} \\
        \toprule
            \multirow{2}{*}{\textbf{Patch Size}}  & \multicolumn{4}{c|}{Text-motion retrieval} & \multicolumn{4}{c}{Motion-text retrieval} \\
             & \small{R@1 $\uparrow$} & \small{R@5 $\uparrow$} & \small{R@10 $\uparrow$}  & \small{MedR $\downarrow$} & \small{R@1 $\uparrow$} & \small{R@5 $\uparrow$} & \small{R@10 $\uparrow$} & \small{MedR $\downarrow$} \\
        \hline\hline
        8$\times$8 & 11.57 & 33.91 & \textbf{50.84} & \textbf{10.00} & 12.20 & 31.83 & 42.88 & 15.00 \\
        \rowcolor[gray]{0.90} 16$\times$16 & 14.02 & \textbf{34.10} & 50.00 & 10.50 & \textbf{13.61} & \textbf{33.33} & \textbf{44.77} & \textbf{13.00}\\
        32$\times$32 & \textbf{14.34} & 33.46 & 48.31 & 11.00 & 12.94 & 32.48 & 42.91 & 14.00 \\
        \bottomrule        
    \end{tabular}   
    }
    \vspace{-5pt}
    \caption{Results of retrieval with different patch sizes.}
    \vspace{-5pt}
    \label{tab:patch_size}
\end{table}

\begin{table}
    \centering
    \setlength{\tabcolsep}{2pt}
    \resizebox{0.99\linewidth}{!}{
    \begin{tabular}{c|cccc|cccc}
        \multicolumn{9}{c}{Dataset: HumanML3D} \\
        \toprule
            \textbf{Training} & \multicolumn{4}{c|}{Text-motion retrieval} & \multicolumn{4}{c}{Motion-text retrieval} \\
            \textbf{Dataset} & \small{R@1 $\uparrow$} & \small{R@5 $\uparrow$} & \small{R@10 $\uparrow$}  & \small{MedR $\downarrow$} & \small{R@1 $\uparrow$} & \small{R@5 $\uparrow$} & \small{R@10 $\uparrow$} & \small{MedR $\downarrow$} \\
        \hline\hline
        \rowcolor[gray]{0.90} HumanML3D & \textbf{10.80} & 26.72 & 38.02 & \textbf{18.00} & 11.25 & \textbf{26.86} & \textbf{37.40} & \textbf{19.50} \\
        Both & 9.99 & 27.22 & 38.64 & \textbf{18.00} & \textbf{11.37} & 25.64 & 36.16 & 21.00 \\
        Both + FT &10.40&\textbf{27.70}&\textbf{38.91}&\textbf{18.00}&11.11&25.86&36.73&20.00\\
        \bottomrule        
    \multicolumn{9}{c}{} \\
    \multicolumn{9}{c}{Dataset: KIT-ML} \\
        \toprule
            \textbf{Training}  & \multicolumn{4}{c|}{Text-motion retrieval} & \multicolumn{4}{c}{Motion-text retrieval} \\
            \textbf{Dataset} & \small{R@1 $\uparrow$} & \small{R@5 $\uparrow$} & \small{R@10 $\uparrow$}  & \small{MedR $\downarrow$} & \small{R@1 $\uparrow$} & \small{R@5 $\uparrow$} & \small{R@10 $\uparrow$} & \small{MedR $\downarrow$} \\
        \hline\hline
        \rowcolor[gray]{0.90} KIT-ML & 14.02 & 34.10 & 50.00 & 10.50 & 13.61 & 33.33 & 44.77 & 13.00\\
         Both &  12.53 & 35.30 & 50.96 & 10.00 & 13.13 & 32.28 & 43.71 & 14.00 \\
         Both + FT & \textbf{17.17}&\textbf{40.46}&\textbf{54.50}&\textbf{8.00}&\textbf{16.76}&\textbf{35.69}&\textbf{46.05}&\textbf{13.00}\\
        \bottomrule        
    \end{tabular}   
    }
    \caption{Results of retrieval with different training datasets. ``Both'' represent the combined datasets of the HumanML3D and KIT-ML datasets. ``Both + FT'' represents the model further fine-tuned on each dataset.
    }
    \vspace{-5pt}
    \label{tab:training_dataset}
\end{table}

\subsection{Training Datasets}
In~\Sref{sec:cross_skeleton}, we demonstrated the effectiveness of our method in cross-skeleton recognition via zero-shot prediction and transfer learning. We further present the results of training our method using a combination of the HumanML3D and KIT-ML datasets in~\Tref{tab:training_dataset}. These results indicate that our method can effectively learn from combined datasets and achieve competitive results on both datasets using a single model. This performance is comparable to the results obtained from separate models trained individually on each dataset. If we further fine-tune the model on each dataset, the proposed method can achieve state-of-the-art performance on the KIT-ML dataset.

\section{Additional Qualitative Results}
\begin{figure*}[t]
    \centering
    \includegraphics[width=1\linewidth]{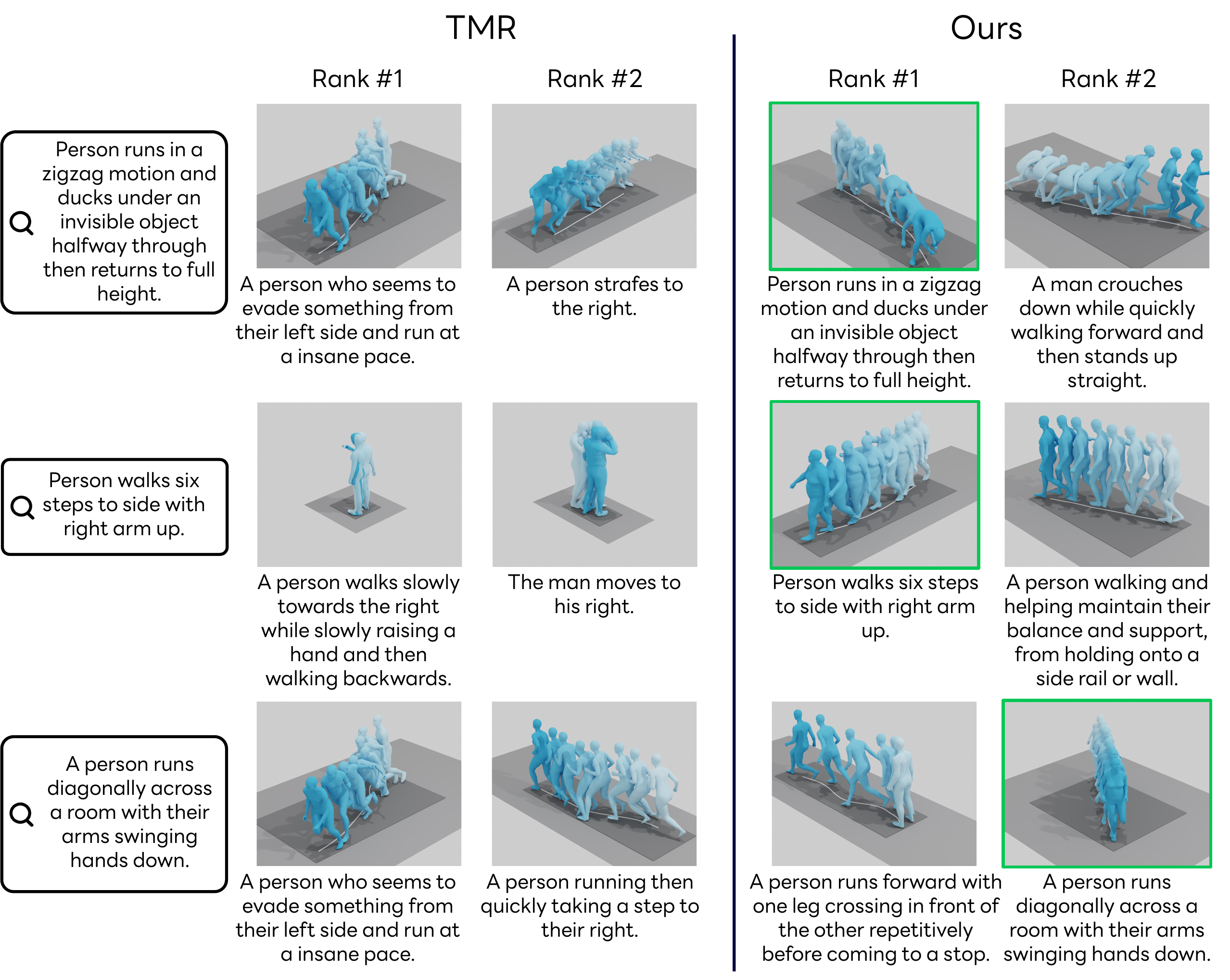}
    \caption{Comparisons of text-to-motion retrieval between TMR~\cite{petrovich2023tmr} and the proposed method. For each query, we show the retrieved motions ranked by text-motion similarity and their accompanying ground-truth text labels. Note that these descriptions are not used in the retrieval process. All motions in the gallery are from the test set and were unseen during training.}
    \label{fig:visualization_vs_t2m}
\end{figure*}

\begin{figure*}[t]
    \centering
    \includegraphics[width=1\linewidth]{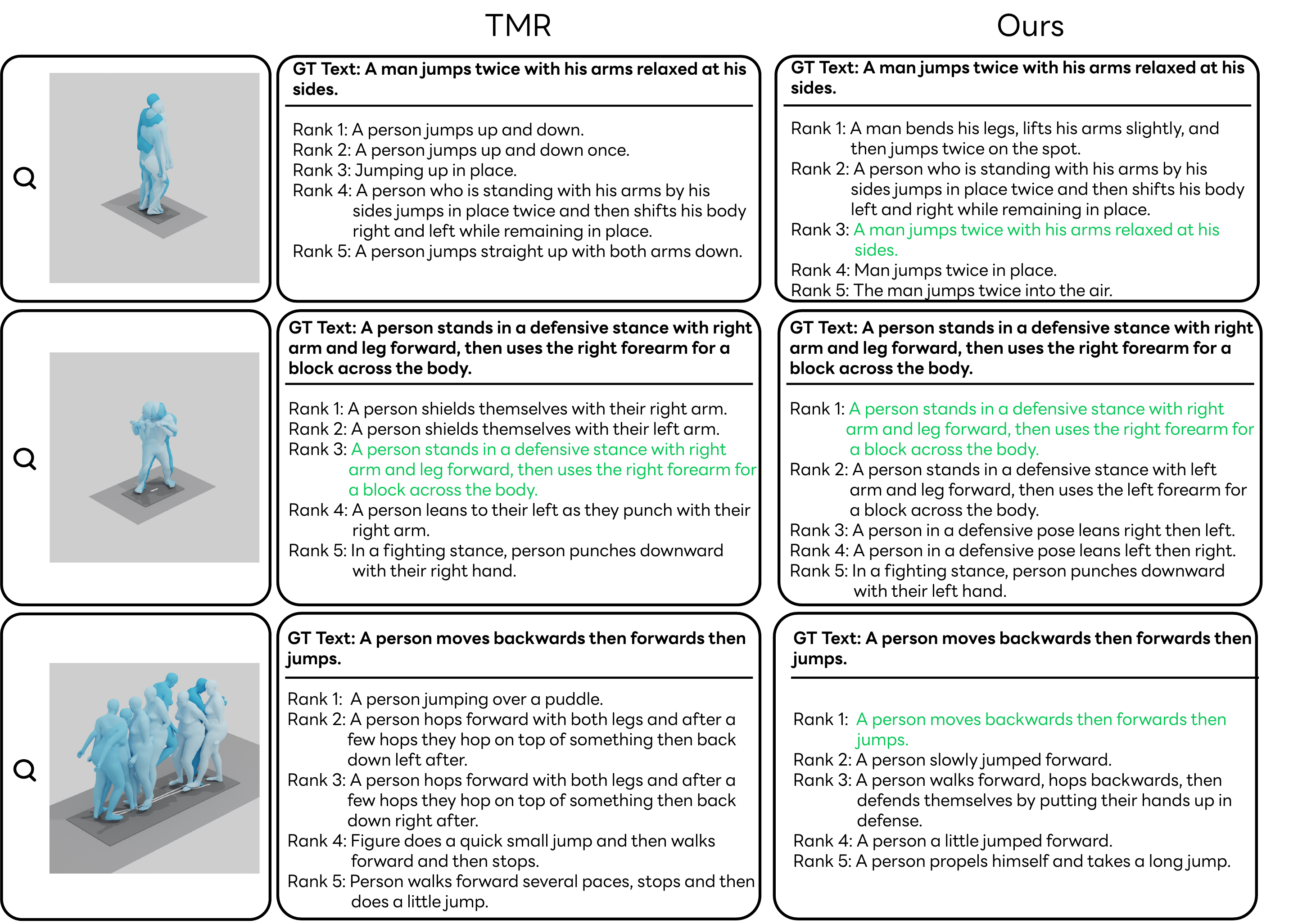}
    \caption{Comparisons of motion-to-text retrieval between TMR~\cite{petrovich2023tmr} and the proposed method. For each query motion, we show the retrieved descriptions ranked by motion-text similarity and their accompanying ground-truth text labels. Note that these ground-truth texts are not used in the retrieval process. All motions in the gallery are from the test set and were unseen during training. For all the samples, our proposed method retrieved reasonable descriptions.}
    \label{fig:visualization_vs_m2t}
\end{figure*}

In this section, we present qualitative results of the text-to-motion retrieval and motion-to-text retrieval tasks with the comparisons between TMR~\cite{petrovich2023tmr} and the proposed method on the challenging HumanML3D dataset. The results of the text-to-motion retrieval are shown in~\Fref{fig:visualization_vs_t2m}. We can find that our method succeeded in finding the motion matching the text descriptions including the details, \eg, ``ducks'' in the first sample and ``with right arm up'' in the second sample. Regarding the motion-to-text retrieval tasks shown in~\Fref{fig:visualization_vs_m2t}, each query motion is displayed on the left, and on the right, we showcase the top-5 retrieved text descriptions along with the ground-truth text labels of query motions. We successfully retrieved the ground-truth descriptions in the top-5 results, and the descriptions in the top-5 results seem to be reasonable to describe the motion sequences except for some mirror-augmented ones. When compared to the results of TMR~\cite{petrovich2023tmr}, our method is better at catching the details of the motion such as ``jumps twice'' in the first sample and ``moves backward then forwards'' in the third sample.

\section{Code}
The code will be released at \url{https://github.com/YU1ut/MotionPatches}. We provide the training codes for building the proposed motion-language model and the test codes for text-to-motion retrieval and motion-to-text retrieval with the HumanML3D and KIT-ML datasets. Please refer to the README in the code repository for details.

\end{document}